\documentclass[letterpaper]{article} 
\usepackage{aaai24}  
\usepackage{times}  
\usepackage{helvet}  
\usepackage{courier}  
\usepackage[hyphens]{url}  
\usepackage{graphicx} 
\usepackage{natbib}  
\usepackage{caption} 
\usepackage{algorithm}
\usepackage{algorithmic}
\usepackage{graphicx}
\usepackage{amsmath}
\usepackage{amssymb}
\usepackage{pifont}
\newcommand{\cmark}{\ding{51}}
\newcommand{\xmark}{\ding{55}}
\usepackage{booktabs}

\usepackage{newfloat}
\usepackage{listings}
\DeclareCaptionStyle{ruled}{labelfont=normalfont,labelsep=colon,strut=off} 
\floatstyle{ruled}
\newfloat{listing}{tb}{lst}{}
\floatname{listing}{Listing}
\title{Playing Atari Space Invaders with Sparse Cosine Optimized Policy Evolution}
\author{
    Jim O'Connor,
    Jay B. Nash,
    Derin Gezgin,
    Gary B. Parker
}
\affiliations{

    Autonomous Agent Learning Lab\\
    Connecticut College\\
    New London, Connecticut, USA\\
    \{joconno2, jnash1, dgezgin, parker\}@conncoll.edu
}

\begin{document}

\maketitle

\begin{abstract}
Evolutionary approaches have previously been shown to be effective learning methods for a diverse set of domains. However, the domain of game-playing poses a particular challenge for evolutionary methods due to the inherently large state space of video games. As the size of the input state expands, the size of the policy must also increase in order to effectively learn the temporal patterns in the game space. Consequently, a larger policy must contain more trainable parameters, exponentially increasing the size of the search space. Any increase in search space is highly problematic for evolutionary methods, as increasing the number of trainable parameters is inversely correlated with convergence speed. To reduce the size of the input space while maintaining a meaningful representation of the original space, we introduce Sparse Cosine Optimized Policy Evolution (SCOPE). SCOPE utilizes the Discrete Cosine Transform (DCT) as a pseudo attention mechanism, transforming an input state into a coefficient matrix. By truncating and applying sparsification to this matrix, we reduce the dimensionality of the input space while retaining the highest energy features of the original input. We demonstrate the effectiveness of SCOPE as the policy for the Atari game Space Invaders. In this task, SCOPE with CMA-ES outperforms evolutionary methods that consider an unmodified input state, such as OpenAI-ES and HyperNEAT. SCOPE also outperforms simple reinforcement learning methods, such as DQN and A3C. SCOPE achieves this result through reducing the input size by 53\% from 33,600 to 15,625 then using a bilinear affine mapping of sparse DCT coefficients to policy actions learned by the CMA-ES algorithm. The results presented in this paper demonstrate that the use of SCOPE allow evolutionary computation to achieve results competitive with reinforcement learning methods and far beyond what previous evolutionary methods have achieved.
\end{abstract}

\section{Introduction}
\label{sec:introduction}

Policy learning for game-playing agents has been extensively explored, with many approaches utilizing large policy networks learned via gradient decent, self-play, neuroevolution, and other methods \cite{csgo,muzero,hneat}. Atari 2600 games and the Atari Learning Environment have developed as a middle ground between traditional board games and modern, complex video games \cite{ale}. A key difference between board games, which are generally dominated by tree searching methods \cite{mcts}, and video games is the presence of noise and irrelevant information. In most cases, this noise must be directly handled by the policy or filtered out by some other method \cite{mnih2015}. Existing noise-filtering techniques have been shown to be effective in a variety of circumstances, but these techniques greatly increase the dimensionality of the overall policy.

While increasing the number of free parameters in a policy is often more effective for gradient-based methods, this generally causes the efficacy of derivative-free optimization methods to decrease. This decrease in efficacy is due to additional free parameters increasing the dimensionality of the search space, and thus increasing the difficulty of optimization without the use of a gradient. In order to avoid increasing the number of free parameters to account for the high dimensional input, SCOPE utilizes the Discrete Cosine Transform (DCT) to compress the input \cite{strang}. The DCT is a domain-agnostic tool that is widely used in applications such as image encoding \cite{DCT,watson}, where it is valued for its ability to concentrate signal energy into a small number of low-frequency components \cite{RAO}.

When applied to neural learning methods, the DCT has been shown to be capable of a type of pseudo-attention as well as showing the ability to augment standard attention methods \cite{Scribano2023,dang,moredct}. Additionally, DCT has been applied to extract additional features from input images to classifier models, which resulted in an increase to classification accuracy on standard benchmarks \cite{lee}. We aim to leverage this pseudo-attention as an all-in-one dimensionality reduction solution, enabling an evolved policy to process a much larger number of inputs than otherwise would be feasible.

In addition to the DCT, SCOPE enforces sparsification in the input coefficient matrix by zeroing out all coefficients below a predefined percentile threshold. This step effectively discards low‐energy components typically corresponding to high‐frequency noise or redundant visual detail, where energy is the absolute magnitude of the coefficient. By retaining only the top‐energy coefficients, we reduce the number of active inputs, which in turn allows the subsequent affine mapping to be realized with fewer weights. Moreover, as the specific positions of the retained coefficients can shift over time, the affine mapping implicitly adapts its receptive “focus” without any explicit attention mechanism, yielding a flexible yet lightweight representation for CMA-ES to optimize \cite{cma}.

\section{Related Work}
\label{sec:relatedWork}

The Atari 2600 platform has served as a benchmark for general AI agents due to its diverse game dynamics and simple interface \cite{ale}. Early work in this area emphasized general game-playing capabilities over domain-specific heuristics. These approaches relied on methods such as linear function approximation with handcrafted features, object-centric representations like Object-Oriented MDPs (OO-MDPs), and planning algorithms that utilized perfect emulators or forward models \cite{mdp}. Feature extraction methods like BASS and DISCO aimed to capture color-based patterns or object classes directly from the screen \cite{bass-disco}, while others leveraged RAM-level observations to model the internal game state.
 
\subsection{Reinforcement Learning Approaches}

\citet{og-dqrl} introduced the first reinforcement learning approaches to ALE through their use of Deep Q-Networks (DQNs) which they used to learn control policies directly from high-dimensional pixel inputs \cite{og-dqrl}. Follow-up work by \citet{req-dqlr} extended this approach by including recurrent layers, attention mechanisms, and \citet{double-ql} introduced Double Q-Learning to address the training instabilities found in previous methods \cite{req-dqlr,double-ql}. Consequently, Distributional Reinforcement Learning (DistRL) was introduced by \citet{dist-rl} as a follow-up to existing DQN-based solutions \cite{dist-rl}. 

Improving upon the success of DistRL in the Atari Learning Environment, \citet{qr-dqn} introduced Quantile Regression DQN (QR-DQN) \cite{qr-dqn}. QR-DQN re-parametrized the distribution of the returns by fixing probabilities and learning value locations through quantile regression. Implicit Quantile Networks (IQN) introduced by \citet{iqn} further extended QR-DQN by learning a continuous quantile function, rather than value locations \cite{iqn}. This improvement allowed IQN to represent a wider family of return distributions, resulting in improved performance across the Atari benchmark.

Following the success of Q-Learning based approaches on the Atari benchmark, \citet{ppo} utilized Proximal Policy Optimization (PPO) as a scalable, robust and simple method for training game agents \cite{ppo, ppo-expand}. PPO based Atari agents achieved performance exceeding that of previous reinforcement learning methods, while being simpler to implement and more compatible with a wide range of architectures. PPO has shown strong performance in both continuous control tasks and Atari games, establishing policy gradient methods as a strong alternative to value-based approaches.

Researchers have also explored how actor-critic methods can benefit from more structured exploration and improved sample efficiency in Atari \cite{a2c}. The Advantage Actor-Critic (A2C) algorithm was developed as a stable alternative to existing Actor-Critic solutions by synchronizing updates across parallel environments. Instead of having multiple agents update the model asynchronously, A2C collects training data from several environments in parallel and performs a single, coordinated gradient update. This approach reduces training noise and leads to more predictable learning dynamics, while maintaining the computational benefits of parallelism.

Recent works continue to push the boundaries of reinforcement learning by enhancing model architecture, learning dynamics, and sampling efficiency. DreamerV2 \cite{dreamerv2} and later DreamerV3 \cite{DreamerV3} demonstrated that world model agents can achieve human-level performance using discrete latent dynamics. Subsequently, \citet{muzero} introduced a major step forward in model-based reinforcement learning through MuZero \cite{muzero}. MuZero combines a learned dynamics model with tree-based planning, achieving state-of-the-art results in Atari. Further work on pushing the boundaries of the ALE benchmark resulted in Agent57 by \citet{agent57}. This solution integrated population-based training, episodic memory, distributed learning, which helped it surpass human-level performance across all 57 Atari games \cite{agent57}.

\subsection{Neuroevolutionary Approaches}
Early research into Neuroevolutionary and biologically grounded approaches in game playing results in HyperNEAT-GGP \cite{hyper-atari} by \citet{hyper-atari}. HyperNEAT-GGP was effective in a limited selection of games such as \textit{Freeway} and \textit{Asterix}, but required the use of domain-specific knowledge to process the input of the Atari system.

Building on the foundational successes of HyperNEAT-GGP, \citet{hneat} applied HyperNEAT to the ALE benchmark and found considerably more success in generalization. These biologically inspired approaches were further refined and improved upon by \citet{dnga} through their introduction of Deep Genetic Algorithms for Atari, where they successfully evolved the weights of a deep neural network based agent.

Following the initial success of evolutionary methods on ALE, OpenAI demonstrated that Evolution Strategies can be a scalable and competitive alternative to standard reinforcement learning algorithms \cite{openai-es} as well. OpenAI-ES achieved strong performance on both MuJoCo and Atari, demonstrating advantages such as the ability to parallelize inherently, robustness to long horizons and sparse rewards, and simplicity of implementation.

Reinforcement learning methods have shown sample efficiency and generalization using deep networks and planning pipelines, while evolutionary strategies offer a strong inherent ability to parallelize and architectural simplicity, but struggle with high-dimensional inputs. SCOPE builds on these two paradigms by introducing a domain-agnostic input compression model that utilizes a bilinear affine mapping that preserves high-utility features via sparsification. This compression enables efficient optimization with CMA-ES, even in tasks typically dominated by reinforcement learning methods.

While more recent methods such as Agent57 \cite{agent57}, Go-Explore \cite{go-explore}, and R2-D2 \cite{r2d2} have set new performance records in the Atari Learning Environment, they do so by leveraging extensive distributed infrastructure, large-scale parallel actors, and training budgets on the order of hundreds of millions to billions of frames. These requirements make them less directly comparable to SCOPE, which is designed as a lightweight, derivative-free approach optimized with fewer than a thousand parameters and trained on a modest computational budget. Our goal is to frame SCOPE not as a replacement for such large-scale systems, but as an alternative that achieves strong results under constrained resources.

\section{The Discrete Cosine Transform}
\label{sec:DCT}

The Discrete Cosine Transform (DCT) is a signal transformation technique that expresses a finite sequence of real numbers as a sum of cosine functions oscillating at varying frequencies \cite{DCT,strang}. Unlike the Fourier transform, which uses complex exponentials, the DCT operates with real-valued basis functions, making it particularly effective in practical applications such as image compression, audio processing, and machine learning \cite{dctreviewimg}.

Among the different variants of the DCT, the type-II DCT is the most commonly used and the one employed in this work. It is especially notable for its energy compaction property; for many natural signals, most of the signal's energy is captured by a few low-frequency coefficients. This makes the type-II DCT well-suited for dimensionality reduction since higher-frequency coefficients, which often correspond to noise or minor variation, can be discarded with minimal information loss.

Formally, for an input vector \( \mathbf{x} = (x_0, x_1, \dots, x_{N-1}) \in \mathbb{R}^N \), the type-II DCT produces a transformed vector \( \mathbf{X} = (X_0, X_1, \dots, X_{N-1}) \), where:
\begin{equation*}
X_k = \alpha_k \sum_{n=0}^{N-1} x_n \cos\left[ \frac{\pi}{N} \left(n + \frac{1}{2} \right) k \right]
\end{equation*}
with normalization constants:
\begin{equation*}
\alpha_k = 
\begin{cases}
\sqrt{1/N}, & \text{if } k = 0, \\
\sqrt{2/N}, & \text{otherwise}.
\end{cases}
\end{equation*}
This normalization ensures orthogonality and energy preservation under the transform.

\begin{figure*}[!t]
    \centering
    \includegraphics[width=0.97\linewidth]{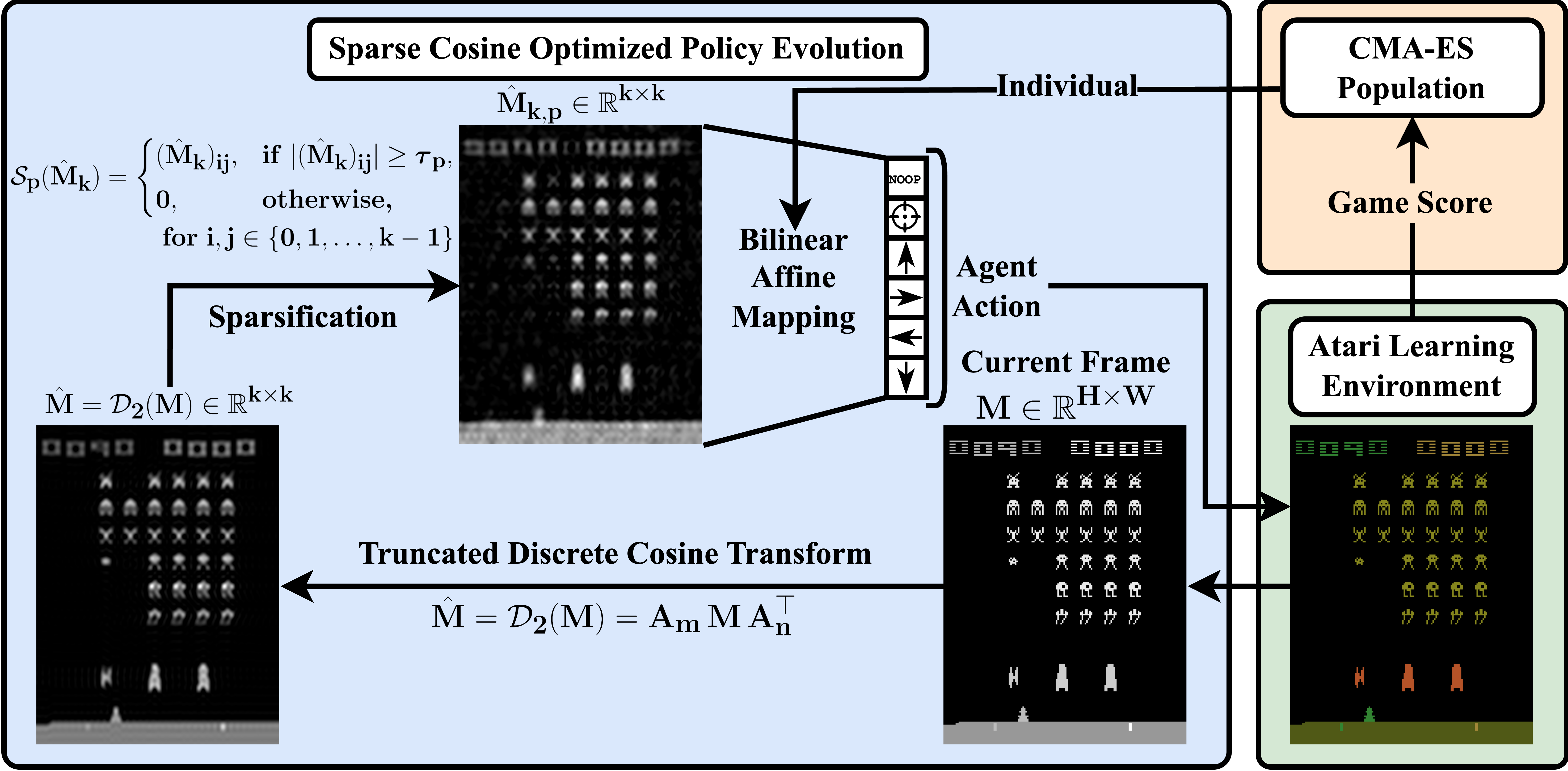}
    \caption{Overview of the SCOPE pipeline. A grayscale Atari frame \( \mathbf{M} \in \mathbb{R}^{H \times W} \) is first transformed via a truncated 2D Discrete Cosine Transform (DCT) into a frequency-domain matrix \( \hat{\mathbf{M}} \in \mathbb{R}^{k \times k} \). A sparsification operator \( \mathcal{S}_p \) retains high-priority coefficients, creating a sparse representation \( \hat{\mathbf{M}}_{k,p} \). This representation is passed through a bilinear affine mapping to produce action logits, which define the agent's behavior in the Atari environment. The Covariance Matrix Adaptation Evolution Strategy (CMA-ES) optimizes the policy parameters and updates the search distribution using episodic game scores.}
    \label{fig:enter-label}
\end{figure*}

SCOPE leverages the two-dimensional version of the type-II DCT for matrix inputs. Given a real matrix \( \mathbf{M} \in \mathbb{R}^{m \times n} \), the 2D DCT is computed by applying the 1D DCT along rows and then columns, resulting in a coefficient matrix \( \hat{\mathbf{M}} \in \mathbb{R}^{m \times n} \):
\begin{equation*}
\hat{\mathbf{M}} = \mathcal{D}_2(\mathbf{M}) = \mathbf{A}_m \mathbf{M} \mathbf{A}_n^\top,
\end{equation*}
where \( \mathbf{A}_m \in \mathbb{R}^{m \times m} \) and \( \mathbf{A}_n \in \mathbb{R}^{n \times n} \) are DCT basis matrices containing type-II DCT basis vectors as rows.

This separable transformation enables a compact and interpretable decomposition of \( \mathbf{M} \). Truncating \( \hat{\mathbf{M}} \) to retain only the top-left (low-frequency) region yields a compressed representation that often preserves the dominant structure of the original signal, while discarding detail and noise.

In this work, the 2D DCT serves as a domain-agnostic method for feature compression. It forms the foundation of SCOPE’s input encoding scheme, enabling scalable learning in high-dimensional state spaces by focusing optimization on the most informative signal components.

\section{Methodology}
\label{sec:methodology}

Sparse Cosine Optimized Policy Evolution (SCOPE) compresses high-dimensional visual input using a frequency-domain transformation and sparsity mask, followed by a learned bilinear affine transformation from the sparse representation to action logits.

While the DCT transformation itself is fully differentiable, the following hard-percentile–based sparsification step introduces a piecewise-discontinuous mapping from input to output, making the overall transformation non-differentiable. This sparsification zeroes out coefficients based on a percentile threshold. Although techniques such as soft-thresholding or straight-through estimators could be applied to approximate gradients, we employ the Covariance Matrix Adaptation Evolution Strategy (CMA-ES), a derivative-free algorithm well-suited for high-dimensional policy search. We formalize each component below.

\subsection{Sparse Cosine Optimized Policy Evolution}
\label{subsec:scope}

Given a grayscale input frame \( \mathbf{M} \in \mathbb{R}^{H \times W} \), SCOPE first applies the 2D type-II Discrete Cosine Transform (DCT-II) to obtain a frequency-domain representation:
\[
\mathbf{\hat{M}} = \mathcal{D}_2(\mathbf{M}) = \mathbf{A}_H \mathbf{M} \mathbf{A}_W^\top,
\]
where \( \mathbf{A}_H \in \mathbb{R}^{H \times H} \) and \( \mathbf{A}_W \in \mathbb{R}^{W \times W} \) are DCT basis matrices.

Dimensionality is reduced by extracting the top-left \( k \times k \) block:
\[
\mathbf{\hat{M}}_k = \begin{bmatrix} 
                        \mathbf{\hat{M}}_{11} & \dots &  \mathbf{\hat{M}}_{1k}\\
                        \vdots & \ddots & \vdots\\
                        \mathbf{\hat{M}}_{k1} &  \dots      & \mathbf{\hat{M}}_{kk} 
                        \end{bmatrix},
\]
which retains the dominant low-frequency components of the input.

To promote selective information flow, a sparsification operator \( \mathcal{S}_p \) is applied to \( \mathbf{\hat{M}}_k \), zeroing out all coefficients below the \( p \)-th percentile in absolute magnitude:
\[
\mathcal{S}_p(\hat{\mathbf{M}}_k)
  = 
  \begin{cases}
    (\hat{\mathbf{M}}_k)_{ij}, 
      & |(\hat{\mathbf{M}}_k)_{ij}|\ge \tau_p,\\[6pt]
    0,           
      & \text{otherwise,}
  \end{cases}
\]

where \( \tau_p \) is the threshold value corresponding to the desired sparsity level. Altogether, SCOPE translates an input image via the pipeline:

\[
  \mathbf{M}
  \;\xrightarrow{\;\mathcal{D}_2\;}\;
  \hat{\mathbf{M}}
  \;\xrightarrow{\;\text{truncate to }k\times k\;}\;
  \hat{\mathbf{M}}_{k}
  \;\xrightarrow{\;\mathcal{S}_p\;}\;
  \hat{\mathbf{M}}_{k,p}
\]

The resulting sparse matrix \( \hat{\mathbf{M}}_{k,p} = \mathcal{S}_p\bigl(\hat{\mathbf{M}}_{k}\bigr)\) serves as input to a bilinear affine policy map, $f_\theta$:
\[
f_\theta(\hat{\mathbf{M}}_{k,p}) = \mathbf{W}_1 \hat{\mathbf{M}}_{k,p} \mathbf{W}_2 + \mathbf{b}=\mathbf{y} \in \mathbb{R}^{m\times n},
\]
where \( \theta = \{\mathbf{W}_1, \mathbf{W}_2, \mathbf{b}\} \), with weight matrices \( \mathbf{W}_1 \in \mathbb{R}^{m \times k} \) and \( \mathbf{W}_2 \in \mathbb{R}^{k \times n} \), and the bias \( \mathbf{b} \in \mathbb{R}^{m \times n} \). This yields an output matrix \( \mathbf{y} \in \mathbb{R}^{m\times n} \), the values of which are flattened and interpreted as unnormalized action logits. In this work, in order to minimize the number of free parameters, we use \( m=1 \) and \( n=6\) to directly map to possible actions.

It is important to note that $f_\theta$ is not parameterized as a neural network. Instead, it is a pure bilinear affine mapping that computes the output by applying two weight matrices to the observation in a linear fashion, followed by a bias term, without any nonlinear activation or deep network structure. This design further reduces the parameter count compared to even the smallest neural network capable of processing the same input--output mapping. For example, the smallest possible neural network to map 10 inputs to 4 outputs would require at least 4 neurons, each with 10 weights and a bias, for a total of 44 weights. In contrast, a bilinear affine mapping only requires a weight matrix of shape $1\times 10$, another of shape $1\times 4$ (adjustable based on the input shape), and 4 bias terms, for a total of only 18 parameters. This significant reduction in free parameters is particularly beneficial for derivative-free optimization, where the search space grows exponentially with parameter dimensionality.

\subsection{Effect of Sparsity on Policy Structure}
\label{subsec:sparsity}

Sparsification introduces a form of conditional computation by defining a partition over the input space. We define the support set
\[
  \eta = \bigl\{(i,j) \in [k]\times[k] \mid [\hat{\mathbf{M}}_{k,p}]_{ij} \neq 0\bigr\},
\] 
i.e., the indices where the matrix is nonzero. Then the space of possible DCT submatrices \( \hat{\mathbf{M}}_{k} \in \mathbb{R}^{k \times k} \) is partitioned into equivalence classes:
\[
\Omega_{\eta} = \{ \hat{\mathbf{M}}_{k} \in \mathbb{R}^{k \times k} \mid  \eta \}.
\]

Each region \( \Omega_{\eta} \) corresponds to a unique sparsity pattern, thus the bilinear affine map \( f_\theta \) behaves as a fixed function restricted by $\Omega_{\eta}$:
\[
f_\theta|_{\Omega_{\eta}}(\hat{\mathbf{M}}_{k}) = \mathbf{W}_1 \Pi_{\eta}(\hat{\mathbf{M}}_{k}) \mathbf{W}_2 + \mathbf{b}
\]

where \( \Pi_{\eta}: \mathbb{R}^{k \times k} \rightarrow V_{\eta} \) is the coordinate projection onto the support \( \eta \), which defines a piecewise-affine function over a partitioned domain, with different sparsity patterns inducing distinct linear regimes.

The number of such regimes is bounded above by \( 2^{k^2} \), corresponding to all possible support patterns, though in practice only a small subset of these are encountered. Because \( \eta \) changes over time depending on the input \( \mathbf{M} \), the active subspace \( V_{\eta}\subset \mathbb{R}^{k \times k} \) is dynamic. The function class induced by SCOPE is thus a union of bilinear affine maps, each operating over a different structured subspace defined by the current input’s cosine coefficient distribution.

Importantly, even though the dimensionality of the input is not reduced, the effective degrees of freedom may be constrained at each time step by the total size of the support matrix. This partition induces a form of implicit regularization and dynamic routing of computation.

\subsection{Covariance Matrix Adaptation Evolution Strategy}
\label{subsec:cmaes}

The Sparse Cosine Optimized Policy Evolution framework (SCOPE) defines a policy \( \pi_\theta \) by applying a sparsified bilinear affine transformation to DCT-encoded inputs. Due to the sparsity applied during the transformation, this function is discontinuous. Thus the parameters \( \theta = \{ \mathbf{W}_1, \mathbf{W}_2, \mathbf{b} \} \) of this transformation cannot be optimized via gradient descent and are instead optimized using the Covariance Matrix Adaptation Evolution Strategy (CMA-ES)~\cite{cma}, a derivative-free stochastic search algorithm suited to high-dimensional, non-convex optimization.

At each optimization step, CMA-ES samples a population of candidate parameter vectors \( \theta_i \sim \mathcal{N}(\boldsymbol{\mu}^{(t)}, \sigma^{(t)^2} \mathbf{C}^{(t)}) \), where \( \boldsymbol{\mu}^{(t)} \) is the current mean, \( \sigma^{(t)} \) is the step-size, and \( \mathbf{C}^{(t)} \) is the covariance matrix capturing dependencies between parameters. Each \( \theta_i \) defines a policy \( \pi_{\theta_i} \), which is evaluated by executing a single episode of the game and recording the final score:
\[
R(\theta_i) = \mathcal{F}(\pi_{\theta_i})
\]

This scalar score is provided by the environment and serves as the fitness function for evolutionary selection. CMA-ES ranks the sampled candidates by their observed final scores and updates \( \boldsymbol{\mu}^{(t)} \) and \( \mathbf{C}^{(t)} \) to increase the likelihood of higher-scoring parameters in subsequent generations.

The transformation from observation to action logits is defined by:
\[
f_\theta(\mathbf{M}) = \mathbf{W}_1 \mathcal{S}_p(\mathcal{D}_2(\mathbf{M})_{[1:k,\;1:k]}) \mathbf{W}_2 + \mathbf{b},
\]
where \( \mathbf{M} \in \mathbb{R}^{H \times W} \) is the raw grayscale input, \( \mathcal{D}_2 \) denotes the 2D DCT-II, and \( \mathcal{S}_p \) is the sparsification operator applied to the top-left \( k \times k \) DCT block. The output of this transformation defines action logits for the agent, which are then used to select actions during gameplay.

Because the support pattern \( \eta \) of the sparse DCT matrix varies across inputs, the effective behavior of the policy is input-dependent, even though \( \theta \) is fixed. Thus, CMA-ES is optimizing a parameterized policy \( f_\theta \) over a distribution of induced subspaces defined by the energy concentration of game observations, measured by the absolute magnitude of cosine coefficients:
\[
J(\theta) = \mathbb{E}_{\mathbf{M} \sim \mathcal{D}_{\text{env}}} \left[ \mathcal{F}(f_\theta(\mathbf{M})) \right]
\]

CMA-ES naturally adapts to this dynamic input structure, concentrating search effort in directions of parameter space that yield robust performance across commonly encountered sparsity patterns. This allows SCOPE to effectively search over policies that generalize across varying visual and structural features in high-dimensional game environments.

\begin{figure*}[!t] 
    \centering
    \includegraphics[width=\linewidth]{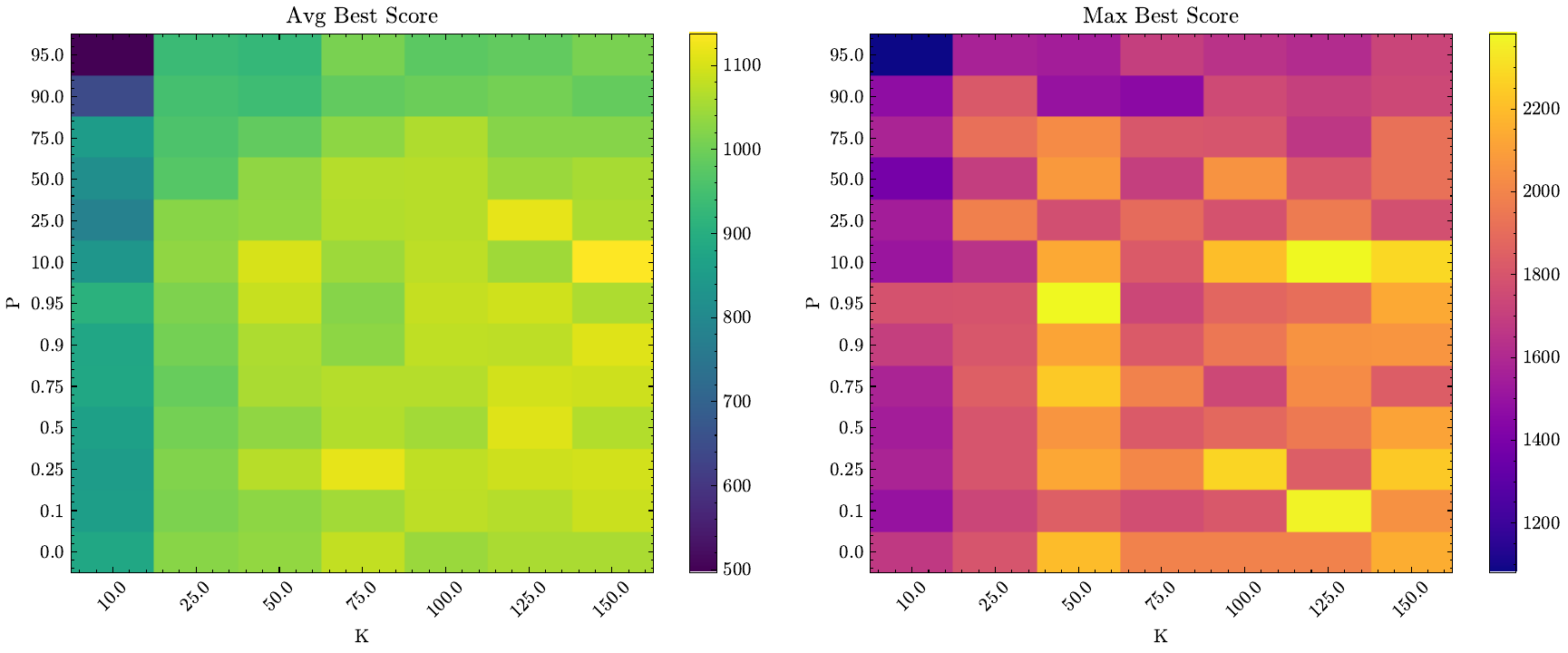}
    \caption{Heatmap of mean final scores over 100 trials with 500 generations each for varying truncation levels $K$ and sparsity percentiles $P$.}
    \label{fig:heatmap}
\end{figure*}

\subsection{Parameter Sweep \& Training Methodology}
\label{subsec:training}

In order to evaluate the effect of changing the level of truncation $K$ and percentile of the enforced sparsity $P$, we then perform a parameter sweep over these hyperparameters. We choose a set of $K$ and $P$ values with $K<min(H,W)$ and $P<100$. We perform a series of short-term trials with each combination, comparing the average and overall best results. The best performing parameter combinations are then chosen for long-term trials with a much higher total number of generations.  

To optimize the SCOPE policy for each combination of parameters, we use the Covariance Matrix Adaptation Evolution Strategy (CMA-ES) in an iterative training loop across a fixed number of generations. In each generation, the algorithm samples a population of candidate policies from a multivariate Gaussian distribution generated by the current covariance matrix. These sampled candidates represent flattened parameter vectors for the bilinear affine mapping used in the SCOPE policy.

For each individual in the sampled population, we evaluate the individuals fitness over 10,000 simulation steps in the Atari Learning Environment or until the end of the episode. The environment is instantiated with deterministic parameters, grayscale observation and a frame skip of 4, consistent with standard Atari settings. During each episode, the SCOPE policy receives a raw frame from the environment, which is passed through the DCT-based compression pipeline that is described in Section~\ref{subsec:scope}. The resulting sparse matrix is then mapped to action logits via the parameterized bilinear affine transformation. The action with the highest logit is selected deterministically at each time step.

The fitness of each individual is defined as the total score of the evaluation. Once all the individuals in the current generation have been evaluated, CMA-ES updates its internal parameters to favor higher-performing solutions. The process is repeated until convergence. 

\section{Arcade Learning Environment}
\label{sec:ale}

The Arcade Learning Environment (ALE) is a benchmark platform for evaluation agents in Atari 2600 games \cite{ale}. It provides a standardized interface that allows an agent to receive visual observations, select from discrete actions, and return episodic rewards. ALE is a testbed for controlled experimentation, allowing configuration of environment parameters such as frame skipping, observations types, and action stochasticity. ALE has become a widely used baseline in artificial intelligence for games research, particularly for comparing the performance of different methods and algorithms \cite{ale, ale2}. The standardized nature of ALE enables consistent evaluation criteria and repeatable results, allowing fair comparisons between methods across numerous Atari games.

In our experiments, we utilize the \texttt{ALE/SpaceInvaders-v5} environment, with a fixed frame skip of 4 and a repeat action probability of 0.0. The environment produces $210 \times 160$ grayscale frames that are used as an input to our SCOPE policy. The action space includes six available actions: \texttt{NOOP}, \texttt{FIRE}, \texttt{LEFT}, \texttt{RIGHT}, \texttt{LEFTFIRE}, and \texttt{RIGHTFIRE}.

\textit{Space Invaders} present a fixed-shooter scenario where the player controls a horizontally mobile spaceship that attempts to destroy horizontally descending enemy spaceships. Points are awarded for shooting enemies, with enemies in higher rows valued higher than lower rows. As the number of enemies decreases, their speed increases, increasing the difficulty of the game. The game terminates when any enemy spaceship reaches the bottom of the screen or when the player dies three times. Space Invaders is a well-fit testbed for evaluating SCOPE due to its structured visual redundancy (i.e. repeating enemy formations), and localized, high-frequency events (i.e. shooting agents, explosions). The pattern-like nature of Space Invaders makes it especially suitable for compression via the Discrete Cosine Transform which transforms an image into low-frequency cosine coefficients to preserve the layout of a scene, while sparsification removes the high-frequency noise without removing the critical signal features. 

In order to test the robustness of the learned SCOPE policies, we test the best policies produced by adding randomness to the environment and measuring the average score attained by the policy. We do this by increasing the repeat action probability to 0.25, which causes each action to have a 25\% chance to be performed twice, as introduced in \citet{ale2}. Any policy that has simply memorized a series of actions that result in a high score would struggle under these conditions, while a reasonable policy would only suffer a partial decrease in efficacy.

To further clarify our experimental setup in terms of determinism and randomness, Table~\ref{tab:randomness} summarizes the key sources of randomness in SCOPE and other methods. Our training phase used deterministic ALE settings (no sticky actions, no random no-op starts), but with the optimizer initialized with a random seed and random weights. To assess generalization and guard against memorization, final trained policies were re-evaluated using a $0.25$ sticky action probability.

\begin{table*}[t]
  \centering
  \fontsize{9pt}{10pt}\selectfont
  \setlength{\tabcolsep}{4pt}

  \caption{Sources of randomness in training and evaluation for SCOPE and baseline methods. OpenAI-ES, DQN, A2C FF, A3C FF (1 Day) referenced from \cite{openai-es}. HyperNEAT referenced from \cite{hneat}.}
  \label{tab:randomness}

  \begin{tabular}{@{}p{5cm}ccccccc@{}}
    \toprule
    Randomness Source & SCOPE & OpenAI-ES & HyperNEAT & A2C FF & A3C FF & DQN & UCT \\
    \midrule
    Optimizer initialized with random seed and random initial weights & \cmark & \cmark & \cmark & \cmark & \cmark & \cmark & N/A \\
    Environment initialized with random number of no-ops at start & \xmark & \cmark & \xmark & \cmark & \cmark & \cmark & N/A \\
    Some level of sticky actions & \cmark\textsuperscript{*} & \xmark & \xmark & \xmark & \xmark & \xmark & N/A \\
    \bottomrule
  \end{tabular}
  \vspace{2pt}

  \footnotesize{\textsuperscript{*}Only applied during post-training evaluation for robustness testing.}
\end{table*}

Running SCOPE on a single core of a 4\,GHz CPU requires approximately 1.5 days for a full training run. This runtime can be substantially reduced through parallelization. For example, when executed on a MacBook Pro equipped with an M2\,Max CPU and 32\,GB of RAM using all available cores, a full run completes in approximately 81 minutes.

\section{Results}
\label{sec:results}

Figure~\ref{fig:heatmap} depicts the outcome of our short-term parameter sweep, illustrating how the truncation parameter $K$ and sparsity threshold $P$ jointly influence SCOPE’s performance. Peak performance is concentrated in the region $K\in[100,150]$ paired with $P\in[5,25]$ with some outliers when $P < 1$. This distribution indicates that moderate frequency truncation combined with low-to-moderate sparsification yields the most effective compression.

Based on these results, we have selected six representative $(K,P)$ configurations for extended evaluation. For each $(K,P)$ configuration, we perform 50 independent training runs. We then report (i) the mean score across all 50 runs, (ii) the mean score of the top-5 runs, 
and (iii) the best single run achieved. Each run is evaluated over a fixed 10{,}000-step episode with no random no-ops at the beginning and a $0.25$ probability of sticky actions, ensuring consistency and robustness in performance assessment.

\begin{table*}[t!]
  \centering
  \fontsize{9pt}{10pt}\selectfont
  \setlength{\tabcolsep}{4pt}

  \caption{Performance of selected SCOPE configurations on \textit{Space Invaders}, reported as in-game score achieved. Policies were trained on a deterministic environment with the best and average scores reported. The final policies were tested in a stochastic version of the same environment to ensure the policies had not memorized.}
  \label{tab:scope-scores}

  \begin{tabular}{@{}lrrrrr@{}}
    \toprule
    SCOPE Parameters & Best & Average ± $\sigma$ & Best (Sto.) & Average ± $\sigma$ (Sto.) & Parameter Count \\
    \midrule
    K=50, P=0.95  & 3,445 & 1,927.4 ± 715.5 & 1,790 & 542.1 ± 300.9 & 350 \\
    K=75, P=0.25  & 3,565 & 1,912.5 ± 707.4 & 2,095 & 599.9 ± 327.7 & 525 \\
    K=125, P=10   & \textbf{4,065} & 1,993.5 ± 672.0 & 1,665 & 464.6 ± 234.5 & 875 \\
    K=125, P=25   & 3,395 & \textbf{2,084.5} ± 595.3 & \textbf{2,280} & \textbf{1,006.1} ± 409.7 & 875 \\
    K=150, P=0.9  & 3,415 & 1,747.1 ± 676.3 & 1,830 & 647.5 ± 349.8 & 1,050 \\
    K=150, P=10   & 3,660 & 2,072.6 ± 724.7 & 2,130 & 931.2 ± 288.8 & 1,050 \\
    \bottomrule
  \end{tabular}
\end{table*}

\begin{table*}[hbt!]
  \centering
  \fontsize{9pt}{10pt}\selectfont
  \setlength{\tabcolsep}{4pt}

  \caption{Comparison of SCOPE configurations and baseline methods on \textit{Space Invaders}. OpenAI-ES, DQN, A2C FF, A3C FF (1 Day) referenced from \cite{openai-es}. HyperNEAT referenced from \cite{hneat}. UCT referenced from \cite{ale}.}
  \label{tab:comparison}

  \begin{tabular}{@{}lrrrrrrrrr@{}}
    \toprule
    Method & K=125,P=10 & K=125,P=25 & K=150,P=10 & OpenAI-ES & HyperNEAT & A2C FF & A3C FF (1 Day) & DQN & UCT \\
    \midrule
    Mean & 1,993.5 & \textbf{2,109.1} & 2,068.3 & 678.5 & 1,251 & 951.9 & -- & 1,449.7 & -- \\
    Top-5 Mean & 3,261 & 3,175 & \textbf{3,347} & -- & -- & -- & 2,214.7 & -- & -- \\
    Best & \textbf{4,065} & 3,395 & 3,660 & -- & 1,481 & -- & -- & -- & 2,718 \\
    Parameter Count & \textbf{875} & \textbf{875} & 1,050 & 680,448 & 116,602 & 680,448 & 680,448 & 1,692,672 & N/A \\
    \bottomrule
  \end{tabular}
\end{table*}

\begin{figure*}[hp!]
    \centering
    \includegraphics[width=\linewidth]{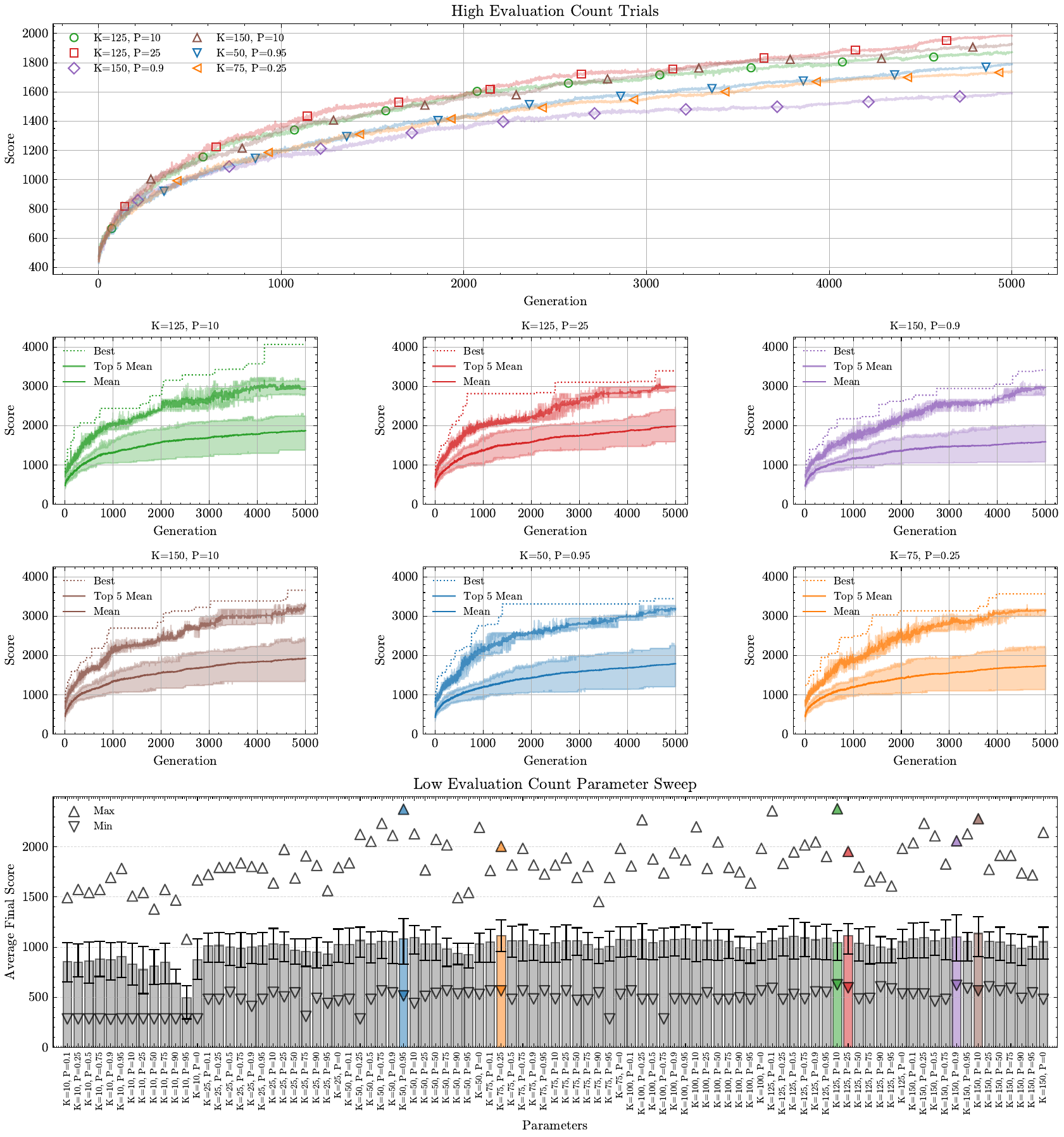}
    \caption{Full accounting of all data gathered via parameter sweep and subsequent trials. The top graph displays the best 6 parameter combinations found by the parameter sweep averaged over 50 trials of 5000 generations for each combination. The center 6 graphs each display a single parameter combination, where each of these graphs displays the mean, 95th percentile, and overall champion over all trials. The lower bar chart displays the full parameter sweep. Each combination is displayed as an average of 100 trials over 500 generations, with the 25th and 75th percentiles displayed as whiskers. The overall maximum and minimum are also displayed, and the parameters chosen for subsequent trials over 5000 generations are highlighted via shading of the min/max indicators. The parameter combination that achieved the highest score is highlighted in green and is displayed in the top-left center graph.}
    \label{fig:appendix_fig}
\end{figure*}

Table~\ref{tab:scope-scores} reports the maximum and mean scores (± standard deviation) achieved over 50 runs of 5000 generations each, under both deterministic conditions and a 25\% action-repeat stochastic setting. The configuration $K=125,P=10$ attained the highest peak score of 4,065, while also maintaining strong average performance. When the final policies were tested with stochastic perturbations, the decline in mean score was limited, demonstrating the robustness of the learned policies. The full accounting of all data collected over both the parameter sweep and subsequent extended evaluation is shown in Figure~\ref{fig:appendix_fig}.

Table~\ref{tab:comparison} compares the best SCOPE policy ($K=125,P=10$) against established evolutionary and reinforcement learning baselines on \textit{Space Invaders}. SCOPE substantially outperforms OpenAI-ES, HyperNEAT, DQN, A2C, A3C (with limited training time), and the UCT baseline provided by ALE despite relying solely on raw pixel inputs and derivative-free optimization.

An important advantage of SCOPE is its significantly reduced parameter count compared to other competitive methods. Table~\ref{tab:comparison} shows that the best-performing SCOPE configuration uses only 875 parameters, significantly fewer than the hundreds of thousands or even millions of parameters reinforcement learning approaches require such as DQN with 1,692,672 parameters.

The substantial reduction in input dimensionality (53\% for $K=125$) coupled with retention of high-energy features enables SCOPE to achieve state-of-the-art performance among derivative-free methods, while also exhibiting resilience to stochastic perturbations. This validates the primary assumption motivating SCOPE, that preserving high-energy DCT coefficients while discarding redundant information strikes an optimal balance between representation compactness and policy expressivity.

\section{Conclusions}
\label{sec:conclusions}

In this work, we have introduced Sparse Cosine Optimized Policy Evolution (SCOPE), a novel framework for evolutionary computation that leverages the Discrete Cosine Transform and percentile‐based sparsification to substantially reduce the dimensionality of Atari game inputs, while retaining the most informative features for policy learning\footnote{The source code of SCOPE can be accessed at \texttt{https://github.com/ConnAALL/SCOPE-for-Atari}}. By embedding this compressed representation within a bilinear affine mapping, SCOPE enables a derivative-free optimizer to approach a high dimensionality pixel-based input with a minimum number of free parameters to optimize.

Our empirical study on \textit{Space Invaders} demonstrates that SCOPE achieves state‐of‐the‐art performance among derivative‐free methods and competes favorably with gradient‐based approaches, all while operating on a reduced input space. The inherent robustness of the resulting policies to action‐repeat stochasticity further highlights SCOPE’s capacity for generalization under environmental variations.

The effectiveness of SCOPE shows the potential for combining signal processing techniques with evolutionary computation methods. This combination not only achieves significant dimensionality reduction but also enhances the interpretability of evolved policies through frequency-domain analysis. This interpretability can open doors to better understanding how evolutionary methods learn in visually complex environments, potentially guiding the development of more explainable and robust AI solutions in game-playing.

Additionally, the use of a bilinear affine mapping as a policy appears to be effective when used with derivative-free optimization methods. As opposed to a linear mapping, a bilinear map is able to accept inputs in matrix form without flattening the input into a vector before computation. This allows the bilinear map to compute the same size input with fewer parameters compared to a linear map. The exact benefit of this trade off between expressivity and parameter count remains to be explored and formalized in future work.

\section{Limitations}
\label{sec:limitations}

While SCOPE shows strong results in the Space Invaders environment, it has several limitations that may impact its generalizability and scalability.

Primarily, all experiments are conducted on a single game. While Space Invaders offers structured visual patterns that align well with frequency-domain compression, it remains unclear whether SCOPE generalizes to environments with less regular spatial structure, more dynamic visual content, or different task objectives.

Second, the current implementation of SCOPE is restricted to discrete action spaces. Extending the method to continuous control tasks would require changes to the output calculation and action selection mechanism, which may not be integrated as naturally with the bilinear affine architecture. Moreover, the compression parameters $K$ and $P$ are treated as constant values which can limit their adaptability in other environments.

Lastly, while the input space is significantly compressed, the bilinear affine mapping still requires a large number of parameters relative to the size of the compressed input. This parameter count can become a bottleneck when scaling to higher resolutions or more complex games.

\section{Future Work}
\label{sec:future}

There are several possible directions for future work regarding SCOPE, several of which we are already investigating. SCOPE’s domain‐agnostic compression mechanism suggests broad applicability to other high‐dimensional control tasks, including continuous‐control benchmarks and three‐dimensional visual domains. In order to apply SCOPE to these domains, we are investigating the possibility of removing the need for a parameter sweep to find adequate $K$ and $P$ parameters. Removing the need for a parameter sweep will greatly increase the efficiency of this algorithm and allow SCOPE to be applied to more complex tasks.

In the near future, we plan to expand our evaluation beyond \textit{Space Invaders} to the full set of 50 widely-tested Atari 2600 games in the Arcade Learning Environment. This broader benchmark will allow us to more thoroughly assess SCOPE’s generalization capabilities across diverse visual structures, dynamics, and reward schemes.

Future work will additionally explore adaptive sparsification and dynamic truncation with the goal of creating a model specifically suited to optimization via derivative-free methods. Specifically, the total information lost by truncation and sparsification can be found via Parseval's theorem \cite{parseval}. Via this theorem, SCOPE could compress an input dynamically until a certain amount of information is lost. This dynamic compression may allow a policy to partition the problem domain based partially on how much a given input is compressed. The effectiveness of such an algorithm is currently under investigation and will be the topic of future work.

\bibliography{references}

\end{document}